\pdfoutput=1

\documentclass[11pt]{article}

\usepackage{emnlp2021}

\usepackage{times}
\usepackage{latexsym}

\usepackage[T1]{fontenc}

\usepackage[utf8]{inputenc}

\usepackage{microtype}

\usepackage{array, multirow, graphicx}
\usepackage{subcaption}

\usepackage{mathtools}
\usepackage{flushend}
\usepackage{cuted}

\usepackage{float}

\usepackage{tablefootnote}
\usepackage{amsmath}

\usepackage{hyperref}
\hypersetup{colorlinks=true,
    citecolor=blue,
    linkcolor=red,
    filecolor=magenta,      
    urlcolor=blue,
}

\title{The Topic Confusion Task: A Novel Evaluation Scenario for Authorship Attribution}

\author{Malik H. Altakrori \\
  School of Computer Science \\
  McGill University / Mila \\
  \texttt{malik.altakrori@mail.}\\  \texttt{mcgill.ca} \And
  Jackie Chi Kit Cheung \\
  School of Computer Science \\
  McGill University / Mila \\
  \texttt{jcheung@cs.} \\ 
  \texttt{mcgill.ca}  
\And
  Benjamin C. M. Fung \\
  School of Information Studies \\
  McGill University / Mila \\
  \texttt{ben.fung@} \\ \texttt{mcgill.ca} \\ 
  }

\date{}

\begin{document}
\maketitle
\begin{abstract}
Authorship attribution is the problem of identifying the most plausible author of an anonymous text from a set of candidate authors. 
Researchers have investigated same-topic and cross-topic scenarios of authorship attribution, which differ according to whether new, unseen topics are used in the testing phase. However, neither scenario allows us to explain whether errors are caused by a failure to capture authorship writing style or by a topic shift.
Motivated by this, we propose the \emph{topic confusion} task where we switch the author-topic configuration between the training and testing sets. This setup allows us to distinguish two types of errors: those caused by the topic shift and those caused by the features' inability to capture the writing styles. 
We show that stylometric features with part-of-speech tags are the least susceptible to topic variations. We further show that combining them with other features leads to significantly lower topic confusion and higher attribution accuracy. Finally, we show that pretrained language models such as BERT and RoBERTa perform poorly on this task and are surpassed by simple features such as word-level $n$-grams.
\end{abstract}

\section{Introduction}
Authorship attribution is the problem of identifying the most plausible author of an anonymous text from a closed set of candidate authors. The importance of this problem is that it can reveal characteristics of an author given a relatively small number of their writing samples. Early approaches to authorship attribution depended on manual inspection of the textual documents to identify the authors' writing patterns, and~\citet{mendenhall1887characteristic} showed that word lengths and frequencies are distinct among authors. 

Since the first computational approach to authorship attribution~\citep{Mosteller.f:1964}, researchers have aimed at finding new sets of features for current domains/languages, adapting existing features to new languages or communication domains, or using new classification techniques, e.g.~\citep{Abbasi.A:2006,Stamatatos.E:2013,silva2011twazn,layton2012authorship,Iqbal.F:2013,zhang2018syntax,malik:2018,Barlas2020}. Alternatively, motivated by the real-life applications of authorship attribution different elements of and constraints on the attribution process have been investigated~\citep{houvardas2006n,Luyckx.K:2011,goldstein2009person,Stamatatos.E:2013,wang-etal-2021-mode}. 

Currently, authorship attribution is being used in criminal investigations where a domain expert would use authorship techniques to help law enforcement identify the most plausible author of an anonymous, threatening text~\citep{Ding.S:2015,rocha2016authorship}. Explaining both authorship attribution techniques and their results is crucial because the outcome of the attribution process could be used as evidence in the courts of law and has to be explained to the jury members.
\begin{figure*}[htb!]
\centering
  \begin{subfigure}{0.33\textwidth}
  \centering
    \includegraphics[trim=20 5 0 0 ,clip,width=.75\linewidth]{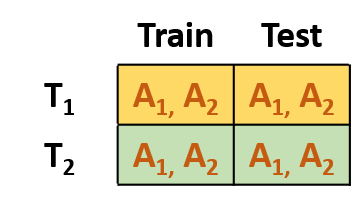}
    \caption{Same-topic\label{Fig:scenarios.a}}
  \end{subfigure}%
  \hfill 
  \begin{subfigure}{0.33\textwidth}
  \centering
    \includegraphics[trim=20 5 0 0 ,clip,width=.75\linewidth]{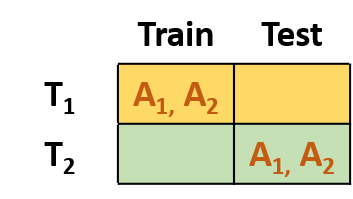}
    \caption{Cross-topic\label{Fig:scenarios.b}}
  \end{subfigure}
  \hfill
  \begin{subfigure}{0.33\textwidth}
  \centering
    \includegraphics[trim=20 5 0 0 ,clip,width=0.75\linewidth]{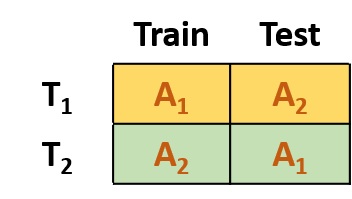}
    \caption{Topic-confusion (Proposed)\label{Fig:scenarios.c}}
  \end{subfigure}%
  \caption{Authorship attribution scenarios. (T: Topic, A: Author)\label{Fig:scenarios}}
\end{figure*}

Researchers have investigated same-topic (Fig.~\ref{Fig:scenarios.a}) and cross-topic (Fig.~\ref{Fig:scenarios.b}) scenarios of authorship attribution, which differ according to whether unseen topics are used in the testing phase. The cross-topic setting is considered more realistic than the same-topic setting, but it causes the performance of well-known authorship attribution techniques to drop drastically. This drop is attributed to the topic-writing style entanglement problem where existing writing style features are capturing the topic variations in the collected documents rather than the authors' writing styles. 

Traditionally, the evaluation of new authorship methods or writing style features for authorship attribution has been based on the difference in the accuracy either on the attribution task or in ablation studies. While this methodology enhanced the performance on the downstream task and helped answer \textit{which} features perform well, there is a need for methods that can help us understand \textit{why} certain features are performing better than others. Specifically, do these newly proposed features/techniques actually capture the stylistic variations of an author, or are they simply better at picking out sub-topic cues that correlate with each author?  %

In this work\footnote{The code will be made available on \url{https://malikaltakrori.github.io/}}, we propose a new evaluation setting, the topic confusion task. We propose to control the topic distribution by making it dependant on the author, switching the topic-author pairs between training and testing. This setup allows us to measure the degree to which certain features are influenced by the topic, as opposed to the author's identity. The intuition is as follows: the more a feature is influenced by the topic of a document to identify its author, the more confusing it will be to the classifier when the topic-author combination is switched, which will lead to worse authorship attribution performance. To better understand the writing style and the capacity of the used features, we use the accuracy and split the error on this task to one portion that is caused by the models' confusion about the topics, and another portion that is caused by the features' inability to capture the authors' writing styles. 

The primary contributions of this work are the following: 

\begin{itemize}
\item We propose topic confusion as a new evaluation setting in authorship attribution and use it to measure the effectiveness of features in the attribution process.
\item Our evaluation shows that word-level $n$-grams can easily outperform pretrained embeddings from BERT and RoBERTa models when used as features for cross-topic authorship attribution. The results also show that a combination of $n$-grams on the part-of-speech (POS) tags and stylometric features, which were outperformed by word- and character-level $n$-grams in earlier work on authorship attribution can indeed enhance cross-topic authorship attribution. Finally, when these features are combined with the current state of the art, we achieve a new, higher accuracy.
\item We present a cleaner, curated, and more balanced version of the Guardian dataset to be used for future work on both same-topic, and cross-topic authorship attribution. The main goal is to prevent any external factors, such as the dataset imbalance, from affecting the attribution results.
\end{itemize}


\section{Related Work}

The first work that used a computational approach is~\citep{Mosteller.f:1964}, which used the Na\"ive Bayes algorithm with the frequency of function words to identify the authors of the Federalist papers~\citep{Juola.p:2008}. Research efforts have aimed at finding new sets of features for current domains/languages, adapting existing features to new languages or media, or using new classification techniques~\citep{f2007identifying,Iqbal.F:2013,Stamatatos.E:2013, Sapkota.U:2014,sapkota2015not,Ding.S:2015,malik:2018}. 

Recent attempts have been made to investigate authorship attribution in realistic scenarios, and many studies have emerged where the constraints differ from the training to the testing samples such as~\citep{bogdanova-lazaridou-2014-cross} on cross-language,~\citep{goldstein2009person,custodio2019ensemble} on cross-domain/genre, and finally,~\citep{sundararajan2018represents,Stamatatos.E:2017,stamatatos2018masking,Barlas2020,Barlas2021} on cross-topic.

\citet{Stamatatos.E:2017, stamatatos2018masking,Barlas2020,Barlas2021} achieved state-of-the-art results on cross-topic authorship attribution. 
\citep{Stamatatos.E:2017, stamatatos2018masking} proposed a character- and word- level $n$-grams approach motivated by text distortion~\citep{granados2012contextual} for topic classification. In contrast to~\citep{granados2012contextual}, ~\citeauthor{stamatatos2018masking} kept the most frequent words and masked the rest of the text.~\citet{Barlas2020,Barlas2021} explored the widely used and massively pretrained transformer-based~\citep{vaswani2017attention} language models for authorship attribution. Specifically, they trained a separate language model for each candidate author with a pretrained embeddings layer from ELMo~\citep{peters-etal-2018-elmo}, BERT~\citep{devlin2019bert}, GPT-2~\citep{radford2019language} and ULMFit~\citep{howard2018universal}. Each model was presented with words from the investigated document, and the most plausible author for that document is the one whose model has the lowest average perplexity.

\begin{figure}[hbt!]
    \begin{subfigure}{0.24\textwidth}
    \centering
    \includegraphics[width=\textwidth]{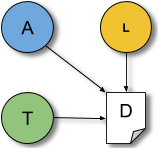}
    \caption{Assumed.\label{causal.perc}}
    \end{subfigure}~
    \begin{subfigure}{0.24\textwidth}
    \centering
    \includegraphics[width=\textwidth]{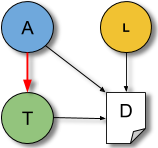}
    \caption{Proposed.\label{causal.prop}}
    \end{subfigure}
  \caption{The relationship diagram between the topic (T), the author's style (A), the language (L), and the document (D).}
  \label{fig:causal}
\end{figure}
\section{The Topic Confusion Task\label{sec:confTask}}
\subsection{Theoretical Motivation}
\begin{figure*}[htb!]
    \centering
    \includegraphics[width=.99\textwidth, trim={0 5.5cm 0 0},clip]{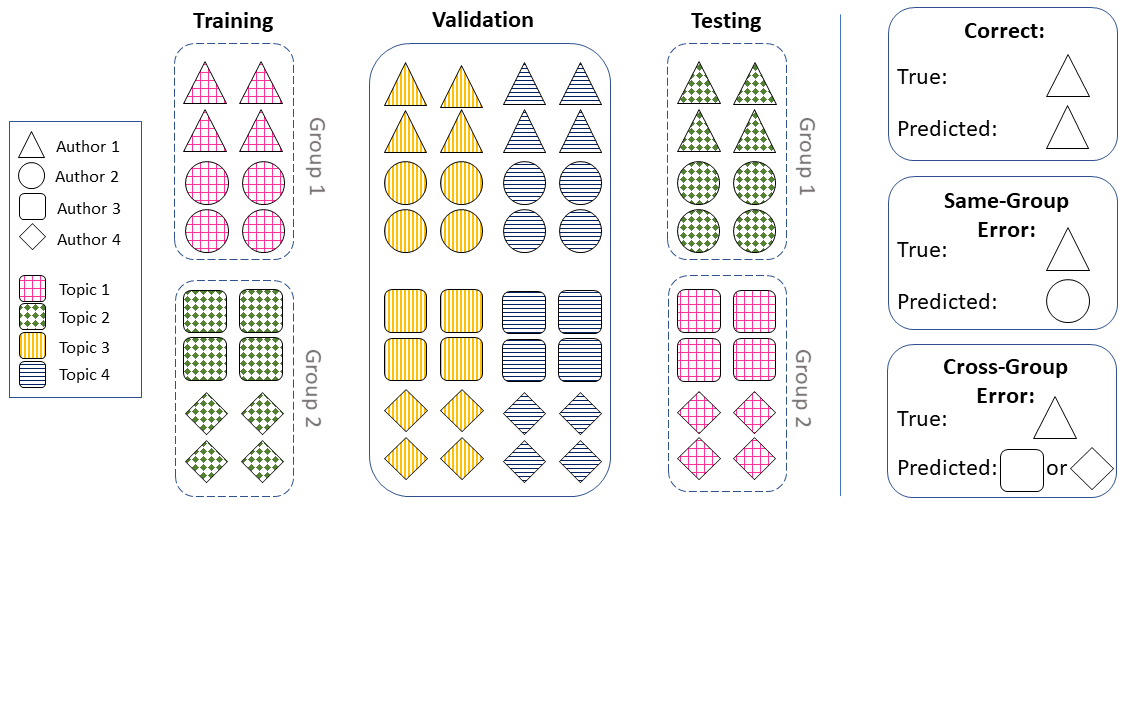}
    \caption{Topic confusion task. We use two topics for training and switch them for testing. Two topics are used for hyperparameter tuning. The topic labels are not available for the classifier during training, and are only used to distribute the samples over the subsets and calculate the scores.}
    \label{fig:topicConf}
\end{figure*}

Figure~\ref{causal.perc} shows the assumed relationship diagram between a document, its author, its topic, and the language rules\footnote{There could be other unknown factors that affect any random variable which the attribution process is not aware of.} that govern the writing process~\citep{Ding.S:2019}. According to~\citet{Ding.S:2019}, these are the factors that affect the process of writing a document. Given a topic's distribution over words, the author picks a subset of these words and connects them using the language rules which govern what words accompany these topical words and how sentences are structured. 

Eq.~\ref{Eq.Joint} shows the joint probability while ignoring the language model, and assuming the topic distribution is independent from that of the author. 
\begin{flalign}
    P(A, T, D) \phantom{xx} &= P(A)P(T)P(D|A,T) \hfill \label{Eq.Joint}\\
    P(A=a|D) \phantom{x}  &\propto \sum_{t}^{T} \left[ 
    P(A=a)P(T=t)\nonumber \right. \\
     & \left. \phantom{xxxxx} P(D|T=t, A=a) \right] \hfill
    \label{Eq.cond}
\end{flalign}
During the attribution process, the model is used to predict an author given an anonymous document using Eq.~\ref{Eq.cond}, which follows from Eq.~\ref{Eq.Joint} after applying Bayes rule. The same argument about the topic also applies to the language model, but for simplicity, we only focus on the topic since POS tags have been shown to capture the stylistic variations in language grammar between authors.

Same-topic scenarios assume that the topic is independent from the author, and that all the topics are available in both training and testing sets. As a result, $T$ in the joint distribution will be set to a fixed value, and $P(T=t)$ is constant $\frac{1}{|T|}$, where $|T|$ is the number of topics in the dataset. If the dataset has only one topic, e.g. $T$= sports then $P(T=sports)$=1 and $P(A=a|D,T)$ is $\propto P(A=a)P(D|A=a)$. This assumption is unrealistic and unintuitive.

In contrast, cross-topic scenarios assume that the topic is independent from the author. This is clear from the cross-topic setup where the topic values are fixed during training and testing. While this setup highlighted a critical flaw in same-topic scenarios and encouraged classification models to rely less on topic cues for authorship attribution, it does not help identify the causes of the errors resulting from changing the topic between training and testing. 

Instead, we propose a setting in which the topic is dependent on the author, as shown in Figure~\ref{causal.prop}, but this dependence varies between training and testing. Our intuition about the effect of the author's writing style on the topic is the following. Consider a topic that has a unique word distribution. When an author writes on this topic, they are bound to generate a slightly different word distribution of that topic in their document. The reason is the limited document length which forces the author to choose a subset of words to describe that specific topic. Now, the topic is dependent on the author's writing choices, and this dependency will vary from one author to another since the same idea can be worded in multiple ways using different word synonyms. 

Because we allow the topic to depend on the author, the joint distribution changes from Eq.~\ref{Eq.Joint} to~Eq.~\ref{Eq.Joint2} and the conditional probability of an author given the anonymous document changes to Eq.~\ref{Eq.cond2}.
\begin{flalign}
    P(A, T, D) \phantom{xx} &= P(A)P(T\textcolor{blue}{|A})P(D|A,T) \hfill \label{Eq.Joint2}\\
    P(A=a|D) \phantom{x} &\propto \sum_t^{T} \left[ P(A=a)  P(T=t\textcolor{blue}{|A=a}) \right. \nonumber \\
    & \left. \phantom{xxxxx} P(D|T=t, A=a) \right]  
    \label{Eq.cond2}
\end{flalign}

Now, we can create a scenario where a learning algorithm only sees samples on one topic for a specific author in the training set but a different topic in the test set, then we measure the error caused by this switch. Note that this proposed scenario will not be as easy as the same-topic, introduces new topics at test time, and can help explain the entanglement of the topic and the writing style.

\subsection{The Proposed Setup\label{subsec:proposed}}
Compared to the standard cross-topic setting, this task can help us understand how a topic affects certain features by showing whether the error is caused by the topic or the features themselves. While the cross-topic setting would give a more realistic performance compared to the same-topic, it lacks any insights on why we got such results.

We propose a new task to measure the performance of authorship attribution techniques given a confounding topic--author setting. The key characteristic of this task is how we associate the topics and the authors in the training, validation and testing sets. Given a set of writing samples written by $N$ authors on $T$ topics where the number of authors $N \geq 4$, the number of topics $T \geq 3$, and each author has, approximately, the same number of writing samples on each topic $T$. 

First, we divide the authors into two equal-sized groups: group 1 and group 2. Next to create the training set, we select two random topics and use writing samples on topic 1 for the authors in group 1 and writing samples on topic 2 for the authors in group 2. For the testing set, we flip the topics configuration that we used for the training set. We use writing samples on topic 2 (instead of 1) for the authors in group 1 and samples on topic 1 (instead of 2) for the authors in group 2. Finally, we use the remaining writing samples on the unused topics for the authors in both groups for the validation set. Figure~\ref{fig:topicConf} shows the setup for the proposed task as an example of having four authors and four topics.

With this setup we can sub-divide the errors that the model makes on the validation and test sets. In particular, we count the following three cases:
\begin{enumerate}

\item \textbf{Correct (\%)}: The ratio of correctly classified samples to the total number of predicted samples.

\item \textbf{Same-group error (\%)}: The number of misclassified samples to authors within the same group as the true author divided by the total number of predicted samples.

\item \textbf{Cross-group error (\%)}: The number of misclassified samples to authors in the other group divided by the total number of predicted samples.
\end{enumerate}

Distinguishing these types of errors allows us to investigate whether features in a classifier tend to be indicative of writing style or topic. In particular, features that are invariant to the topic and only capture the authors' writing styles should lead a model to correctly identify the author in the test set. Conversely, features that capture the topic instead of the writing style would lead a model to classify according to topic, resulting in cross-group errors. Finally, a model that fails for other reasons---either because the writing styles are too similar or because the used features can only partially capture the writing styles---will misclassify samples to authors within the same group.

\section{Dataset\label{subsec:dataset}}
We present an extended, curated, and relatively balanced version of the Guardian dataset.\footnote{Appendix~\ref{dataColl} describes the data collection procedure.}
One motivation is that the articles in the commonly used version of the dataset contained some HTML artifacts and meta-data from the Guardian's website, and had a number of its articles either on the wrong topic, or written by authors that are not in the dataset. Because of that, we retrieved the original articles, and added more articles to balance the number of writing samples per author on each topic. We maintained the same upper limit on the number of documents per author as the original dataset.

Another reason is that as we try to understand the effect of the topic on the attribution process, we need to isolate any external factors that may affect the performance and make the results noisy. For example, in the topic confusion task, we have to use topics with writing samples from all the authors. Otherwise, the model could learn to favour one topic versus the other during training, while on test time, it will have author samples that it did not see during training. Based on that, it will be hard to tell whether these samples will be misclassified due to lack of training samples or due to a strong topic effect on the attribution process. Although datasets in real life can be imbalanced, this issue can be addressed by randomly excluding some writing samples to make the dataset imbalanced or using proper performance metrics for imbalanced datasets such as weighted accuracy, precision, recall and F-Score. The number of collected articles and additional descriptive statistics are provided in Table~\ref{tab:myData}.

\begin{table*}[ht]
    \centering
    \begin{tabular}{ll |lll }
    \hline 
    \textbf{Total number of:} & &
    \multicolumn{2}{l}{\textbf{Number of articles per topic}} &
    \\\hline
    \phantom{Hi} Topics     & 4           & \phantom{Hi} Politics (P)  & 130            \\ 
    \phantom{Hi} Authors    & 13          & \phantom{Hi} Society (S)  & 118*          \\  
    \phantom{Hi} Articles   & 508         & \phantom{Hi} UK (U)      & 130           \\
    \phantom{Hi} Words      & 3,125,347   & \phantom{Hi} World (W)   & 130            \\
     \hline
     \textbf{Average number of: } & &
     \multicolumn{2}{l}{\textbf{Number of articles per author}} \\\hline
     \phantom{Hi} Articles / Author   & = 39.1 ($SD=1.5$)                         & \phantom{Hi} M.K.                & 35\\
     \phantom{Hi} Articles / Topic    & = 127 ($SD=5.2$)                          & \phantom{Hi} H.Y.                & 37\\
     \phantom{Hi} Words / Author      & $\approx$ 41 K ($SD$ $\approx$ 6.9 K)   & \phantom{Hi} J.F.                & 38\\
     \phantom{Hi} Words / Topic       & $\approx$ 781 K($SD$ $\approx$ 13.0 K)  & \phantom{Hi} M.R. and P.P.       & 39 \\
     \phantom{Hi} Words / Document    & $\approx$ 1050.2                        & \phantom{Hi} \textbf{The remaining 8}   & 40 \\ \hline
     \end{tabular}
    \caption{Descriptive statistics for the extended Guardian dataset (* Has less than 10 articles per author).}
    \label{tab:myData}
\end{table*} 

\section{Authorship Attribution Models\label{subsec:QWS}}
In this section, we discuss two groups of authorship attribution models. The first group contains a set of classical models that use hand-engineered features and a classification algorithm. The second group comprises a set of neurally-inspired models motivated by recent advancements in many natural language processing tasks. Such models are considered end-to-end system where the feature representation is learned by the model as opposed to being hand-crafted and provided to the model. 

\subsection{Classical Features with SVM} 
This approach uses a set of classical, hand-engineered features with a non-neural classification algorithm. We experiment with a wide spectrum of features that include both stylometric features and $n$-gram features. Early work on authorship attribution proposed using stylometric features to represent an author's writing Style. On the other hand, $n$-gram features were used with most text classification tasks until recent neural representations replaced them.

With all the following features, we used the instance-based approach~\citep{Stamatatos.e:2009} where a writing style is extracted from every sample separately. A classification model is trained on the extracted features to predict the authors of new, unseen samples. We used~\citet{scikit-learn}'s implementation of linear Support Vector Machines (SVM) as the classification algorithm\footnote{Appendix~\ref{appOptimal}, Tables~\ref{hypers} and~\ref{optimalTable} show the range of values and the average optimal parameters that are fine-tuned on the validation set, respectively.}, which is a common choice in authorship attribution~\citep{Stamatatos.E:2017}. 

Different classification algorithms can be used with these features. Examples are Na\"ive Bayes, decision trees and SVM. We chose to use SVM with linear kernel based on its favorable performance in previous work~\citep{Sapkota.U:2014,sapkota2015not,Ding.S:2015,Stamatatos.E:2017,stamatatos2018masking}. 

\paragraph{Stylometric Features~\citep{Iqbal.F:2008,Iqbal.F:2013}.}
We evaluate 371 features including syntactic features and lexical features on both character- and word-level. These features are listed in Appendix~\ref{styloFeat}-Table~\ref{tbl:features}.

\paragraph{Character-, Word- and POS-level N-Grams~\citep{Stamatatos.E:2013,Sapkota.U:2014,sapkota2015not}.}
Using $n$-grams is a common approach to represent documents in authorship attribution. In most text classification tasks, the tokenization is done on either the word or the character level. We use both character and word level $n$-grams in addition to POS-level~\footnote{We used the POS tagger from~\citep{manning2014stanford}.} $n$-grams which are proven to be an essential indication of style~\citep{Ding.S:2015,sundararajan2018represents}. 

\paragraph{Masking~\citep{Stamatatos.E:2017,stamatatos2018masking}.}
This preprocessing technique replaces every character in words to be masked with a (*) and replaces all the digits with a ($\#$). Masked words are chosen based on their frequency in the British National Corpus (BNC), an external dataset. After Masking, tokens are put back together to recreate the original document structure before extracting $n$-gram features. 

\paragraph{Combining features.} One advantage to using hand-engineered features on the sample level is that these features can easily be combined. First, we evaluated the combination of the stylometric features and POS $n$-grams. Next, we combined both these features to the other classical features mentioned above. 

\subsection{Pretrained Language Models}

\paragraph{Few-Shot BERT and RoBERTa.}
This is an example of a few-shot classification with pretrained language models. We used a sequence classification model with a pretrained embeddings layer from the transformer-based non-autoregressive contextual language models BERT~\citet{devlin2019bert} and RoBERTa~\citep{liu2019roberta} followed by a pooling layer then a classification layer. Given the huge size of these models and the small number of training samples, we decided to freeze the embeddings and train only the classification layer. We used the implementation provided by the HuggingFace~\citep{Wolf2019HuggingFacesTS} library\footnote{\url{https://huggingface.co}}.

\paragraph{Author Profile (AP) BERT and RoBERTa~\citep{Barlas2020,Barlas2021}.} 
We trained a separate neural language model for each author in the dataset where the embedding layer is initialized with embeddings from BERT and RoBERTa. To predict the author, we used each language model --or author profile-- to calculate the average perplexity of the model for an investigated document. Before attribution, however, the perplexity scores are normalized using a normalization vector ($n$) to make up for the biases in the output layer of each language model, where $n_i$ equals the average perplexity of profile $A_i$ on the normalization corpus.

\citet{Barlas2020,Barlas2021} used two normalization corpora during inference: the training set (K) and the testing set without labels (U). The author with the lowest normalized perplexity score is the most plausible author of the investigated document. Note that assuming the availability of a test set rather than a single document is unrealistic in authorship attribution even if labels were not provided. We evaluated both cases for the sake of completeness.  

\begin{table*}[!htb]
\centering
\caption{\label{tbl:topicConf} Average results (SD) on the topic confusion task and the cross-topic scenario. The last row is random performance. (\textbf{Boldface}: Best result per column. $\uparrow$ Higher is better. $\downarrow$ Lower is better. \%: Percentage. \textcolor{red}{$^{*}$}State of the art. \textcolor{red}{$^{**}$} Has access to the (unlabeled) test set.)
}
\begin{tabular}{l|ccc|||c}
\hline
& \multicolumn{3}{c|||}{\textbf{Topic Confusion}}& \textbf{Cross-topic}    \\\hline
\multirow{2}{*}{\textbf{Models}}  & $\uparrow$ \textbf{Correct} & $\downarrow$ \textbf{Same-group} & $\downarrow$ \textbf{Cross-group} & \textbf{$\uparrow$ Accuracy} \\
 &  & \textbf{Error} & \textbf{Error} &   \\\hline\hline 

Stylo.                      & 63.1 (4.2)  & 15.7 (2.7)  & 21.2 (3.0)  &  61.2 (3.1) \\\hline
POS $n$-grams               & 72.0 (4.5)  & 11.5 (2.9)  & 16.6 (3.3)  &  71.0 (3.2) \\
\phantom{hi} + Stylo        & 79.6 (4.0)  & \phantom{0}8.4 (2.6)  & 12.1 (2.8)  &  79.2 (2.7) \\\hline

Char $n$-grams              & 70.1 (6.5)  & \phantom{0}6.8 (2.4)  & 23.2 (6.5)  &  77.3 (2.8) \\			
\phantom{hi}+ Stylo         & 73.0 (6.4)  & \phantom{0}6.5 (2.6)  & 20.5 (6.1)  &  - \\
\phantom{hi}+ Stylo \& POS  & 76.8 (6.1)  & \phantom{0}\textbf{6.0} (2.3)  & 17.2 (5.6)  &  82.8 (2.7)\\\hline

Word $n$-grams              & 62.5 (7.4)  & \phantom{0}7.9 (2.7)  & 29.6 (7.4)  &  77.7 (2.7\\
\phantom{hi}+ Stylo         & 72.4 (6.4)  & \phantom{0}7.3 (2.3)  & 20.3 (6.2)  &  - \\
\phantom{hi}+ Stylo \& POS  & 80.3 (5.0)  & \phantom{0}7.1 (2.7)  & 12.6 (4.2)  &  \textbf{83.3} (2.6)\\\hline

Masking (Ch.) \textcolor{red}{$^{*}$}               & 79.5 (5.6)  & \phantom{0}6.8 (2.7)  & 13.8 (5.0)  & 80.9 (2.6) \\
\phantom{hi}+ Stylo \& POS  & 83.1 (4.8)  & \phantom{0}6.4 (2.7)  & 10.4 (3.5)  & 83.2 (3.3)\\\hline

Masking (W.)                & 76.8 (5.7)  & \phantom{0}7.9 (2.9)  & 15.3 (5.7)  & 77.9 (4.0) \\
\phantom{hi}+ Stylo \& POS  & \textbf{83.3} (4.4)  & \phantom{0}6.7 (2.7)  & \textbf{10.0} (3.2)  & 82.8 (3.3) \\ \hline \hline

FS BERT                     & 33.1 (5.7)  & 19.9 (5.6)  & 47.0 (9.0)  & 37.5 (3.5)\\
BERT AP (K)                 & 51.6 (7.5)  & \phantom{0}8.2 (3.1)  & 40.2 (8.6)  & 67.3 (4.4)\\
BERT AP (U)                 & 52.1 (7.3)  & \phantom{0}8.4 (3.2)  & 39.6 (8.5)  & 71.1 (3.3)\\ \hline

FS RoBERTa                  & 39.8 (7.5)  & 13.1 (5.1)  & 47.1 (10.9) &  51.1 (3.4)\\
RoBERTa AP (K)              & 57.8 (7.1)  & \phantom{0}7.1 (2.9)  & 35.1 (8.5)  &  70.8 (2.0)\\
RoBERTa AP (U\textcolor{red}{$^{**}$})              & 58.9 (7.1)  & \phantom{0}6.8 (2.8)  & 34.3 (8.3)  &  75.8 (3.8)\\

\hline\hline
 \multirow{1}{*}{\textit{``random chance"}} & 8.3 & 41.7 & 50.0 & 7.7\\\hline
\end{tabular}

\end{table*}

\section{\label{subsec:eval}Evaluation Procedure}
For each set of features, we used the setup explained in Section~\ref{sec:confTask} to create a 100 different configurations. For each configuration, we randomly ordered the topics, selected 12 out of the 13 available authors, and distributed the authors to the groups. This setting is considered as one single experiment. To account for randomness in the classification algorithm, we repeated every single experiment ten times\footnote{We trained FS BERT and FS RoBERTa only once.}, and reported the average balanced accuracy score and standard deviation.

We decided to omit one author and use the remaining twelve out of the available 13 authors to balance the groups. With this split, the probability of picking the correct author is $\frac{1}{12}$, the likelihood of choosing a wrong author in the same group is $\frac{5}{12}$, and the probability of picking a wrong author in the other group is $\frac{6}{12}$. This case applies if the true author was in either group 1 or group 2. However, suppose we were to use all the 13 authors and divide them into two groups of six and seven authors, respectively. In that case, the probabilities will differ depending on whether the actual author is in the group with six authors or seven authors. In that case, we will need to re-weight the errors based on their probability, and that will complicate the results as we will not be talking about the exact number of samples.  

After creating the training, validation, and testing sets we train models for authorship attribution. First, the features are extracted from the writing samples. Second, a classification model is trained on the training samples, tuned on the validation set to pick the best hyperparameters, and tested on the testing set. Note that the classifier does not have access to any information about the setup, such as the groups configuration or the topic labels. 

\section{Results and Discussion}
\subsection{Topic Confusion Task}
Table~\ref{tbl:topicConf} shows the results on the topic confusion task using the proposed measured in section~\ref{subsec:proposed}. Correct is the percentage of samples that were correctly classified, same-group error is the percentage of samples that were attributed to the wrong author but within the same group as the correct author, and finally cross-group error is the percentage of samples that were attributed to the wrong author and to the author group that does not contain the correct author ---caused by the change in the topic---. 

\paragraph{Classical Features with SVM.}
Compared to stylometric features, a classifier using character $n$-grams would correctly classify more samples. However, splitting the error shows that using stylometric features will lead to a lower cross-group error, which is associated with the topic shift. Here, the topic shift does not cause the low performance of stylometric features but rather because they partially capture the writing style.

When looking at character- vs word-level $n$-grams, we see that they have comparable same-group errors while cross-group error is much higher for word $n$-grams. Our results are in line with the literature on the classical cross-topic authorship scenario, which shows that character $n$-grams outperform word $n$-grams while still capturing the topic, which makes character $n$-grams less influenced by the topic in the attribution task.

Next, we look at the effect of masking as a preprocessing technique. Specifically, we compare character- and word-level $n$-gram features before and after masking. Masking infrequent words is evident in the cross-group error between character $n$-grams and masking on the character-level as well as the word $n$-grams and masking on the word level. Table~\ref{tbl:topicConf} shows the same-group error remained fixed while cross-group error decreased by around 10\% and 15\% for the character- and the word-level, respectively. 

\paragraph{Combining features.}
We evaluated the effect of combining both stylometric features and POS $n$-grams with character- and word-level $n$-grams with and without masking. The results of combining both stylometric features and POS $n$-grams with all the other features have decreased the cross-group error significantly, which resembles less confusion over the topic. On the other hand, the same-group error was reduced by merely one sample at max in most cases. 

\paragraph{Pretrained Language Models.} Surprisingly, such models performed very poorly on this topic confusion task regardless of the attribution approach being used with them. According to the results, these models have a much larger cross-group error which is associated with the topic shift. 

One potential explanation for this behavior is that in authorship attribution, two words would have similar embeddings if they appear in a similar context \textit{and} are used by the author in their writing samples. Consider the words `color' and `colour' for example. These are essentially the same word but with different spelling based on whether American or British English is being used. Ideally, these two words would have very similar embeddings, if not identical ones. The distinction between the two is critical because it indicates the author's identity or the language system they use. Authorship attribution techniques highlight these differences and use them to identify the most plausible author of an anonymous document. 

\subsection{Comparing the Performance on the Cross-Topic Scenario}
We use the cross-topic scenario on the Guardian dataset to compare the performance of different attribution models to that on the topic-confusion task. Note that it is common to do the evaluation on one of the two cross-topic authorship attribution datasets~\citep{goldstein2009person,Stamatatos.E:2013} similar to~ \citep{goldstein2009person,Sapkota.U:2014,Stamatatos.E:2017, stamatatos2018masking,Barlas2020,Barlas2021}

The last column of Table~\ref{tbl:topicConf} shows a similar trend to the topic-confusion task where combining stylometric features and POS-level $n$-grams to other classical features results in better authorship attribution. Notably, the combination of stylometric features, POS- and word-level $n$-grams outperforms the state-of-the-art. Additionally, adding stylometric features and POS-level $n$-grams to masking (Ch) and masking (W) achieved better performance than state of the art, but the difference was statistically significant only when we comparing with masking (Ch.) ($P=0.04$). Appendix~\ref{additional} contains the experimental setup, detailed results and analysis supported with statistical significance tests and an ablation study on the cross-topic scenario.

Finally, consider two completely different approaches to authorship attribution, namely BERT AP (U) and a linear SVM with POS-level $n$-grams. Now, note how the accuracy alone on the cross-topic scenario does not provide any insights on why these two models perform very similarly. In contrast, the cross-group error in the topic-confusion task shows that a linear SVM with POS-level $n$-grams has a much lower error, hence, less affected by the change in topic compared to BERT AP (U).

\section{Conclusion}
In this work, we proposed the topic confusion task, which helps us characterize the errors made by the authorship attribution models with respect to the topic. Additionally, it could help in understanding the cause of the errors in authorship attribution. We verified the outcomes of this task on the cross-topic authorship attribution scenario. We showed that a simple linear classifier with stylometric features and POS tags could improve the authorship attribution performance compared to the commonly used $n$-grams. We achieved a new state-of-the-art of 83.3\% on the cross-topic scenario by resurrecting stylometric features and combining them with POS tags and word-level $n$-grams, 3\% over the previous state-of-the-art, masking-based, character-level approach. Surprisingly, neurally-inspired techniques did not perform well on the authorship attribution task. Instead, they were outperformed by a simple, hand-crafted set of stylometric features and POS-level $n$-grams and an SVM classifier with a linear kernel. 

\section*{Acknowledgments}
This work was supported by the Doctoral Scholarship from Fonds de Recherche du Quebec Nature et Technologies (FRQNT-275545), Discovery Grants (RGPIN-2018-03872) from the Natural Sciences and Engineering Research Council of Canada, and Canada Research Chairs Program (950-232791). The second author is supported by a Canada CIFAR AI Chair.

\section{Ethics/Broader Impact Statement}
\noindent$\bullet$ The \textbf{data collection} process is described in Appendix~\ref{dataColl}. Scripts to retrieve the articles are also provided in the supplementary material. As per the requirements of the Guardian API, we do not share the actual articles, but rather the URLs and the script to extract the original articles. In addition to ownership rights, this is important in case the original authors of these articles decide to delete them, or make some modifications to these articles. The articles that we use are available online, and do not discuss sensitive topics that, if shared, could hurt the original authors of these articles. 

\noindent$\bullet$ All the \textbf{manual work} that was required, such as removing the authors names from the body of the articles, was done solely by the authors. No external/paid help was required. Details of all the manual work is provided in the supplementary material (not the Appendix). To ensure reproduciblity of this manual work, we provided a list of all the steps and how to perform them. 

\noindent$\bullet$ To ensure the \textbf{quality of the dataset}, we manually inspected the documents for potential features that would reveal the identity of the authors easily. The dataset has 508 documents which makes the task of manually inspecting the documents tedious, but possible. Future work does not need to do this inspection. We provide a list of all the required manual changes in the supplementary material. 

\noindent$\bullet$ \textbf{The intended use of this work.} One application of authorship attribution is in crime investigation where a domain expert can help law enforcement identify the true author of an anonymous investigated text. This anonymous text can be a threatening message, or a suicide note. In the famous case of Ted Kaczynski, also known as the ``Unabomber", linguistic evidence was used to identify Kaczynski by comparing his PhD thesis to the communication letters and the ``manifesto" sent to the investigating authorities by the Unabomber. 

Another area that could benefit from this work is research on anonymization, which is the task of hiding the identity of an author of a document to protect their privacy. To evaluate anonymization techniques, their outcome, i.e., the anonymized documents, are presented to an authorship attribution technique to identify their original author after anonymization. The effectiveness of these anonymization techniques is based on the change in the attribution accuracy before and after anonymization. As this work aims to provide a better understanding of what makes a writing style, we hope that this would lead to better anonymization techniques. 

Finally, it is important for the public to know about the existence of such authorship techniques which can identify them using small number of their writing samples. They need to know that their identities are not completely protected by the anonymity of the internet. If a small research group can develop such techniques, then governments and organizations with more budget and personnel can do the same or more, if they intend to. 

\noindent$\bullet$ \textbf{Failure mode} is exactly what we try to address in this work. Current authorship attribution techniques are highly affected by the topic of the documents, hence the outcome of the attribution process could potentially pick the wrong candidate due to topic similarity between the author's writing samples and the investigated document, and not the actual writing style. 

\noindent$\bullet$ \textbf{Potential misuse} can occur if this work is used against people who benefit from the internet anonymity to express their opinions against oppressing governments or individuals. In this case, individuals or governments would use the same techniques to identify people who speak against their interests, and persecute them. 

\newpage
\bibliography{main}
\bibliographystyle{acl_natbib}

\appendix

\section*{Appendices}
\section{Data Collection\label{dataColl}}
First, we curated the existing dataset by retrieving the 381 original documents from the Guardian's website. Next, we inspected the authors' names and the topics associated with each article. We excluded the articles that had the wrong topic (e.g. labelled as ``Politics" in the dataset while having a ``Society" tag on the website), or the ones that appeared under more than one of the previous topics, or were co-authored by multiple authors. 

Next, we used the Guardian's API\footnote{\url{https://open-platform.theguardian.com}} to get all the articles written by each author, filtered them based on the topic, and collected the URLs of these articles and new articles aiming for 10 documents per author per topic. This resulted in a total of 40 documents per author. Note that while some authors have been writing in the Guardian for more than 20 years, they would mostly focus on one topic while occasionally writing on the other four. As a result, we still could not get 10 articles per author on the Society topic. The supplementary material contains full instructions, and the necessary script to get the data and preprocess it.

\section{Stylometric Features\label{styloFeat}}
See Table~\ref{tbl:features} below. 

\begin{table*}[htbp]
\centering
\begin{tabular}{p{7cm}}
\hline
\textbf{Lexical Features - Character-Level}  \\\hline
1. Characters count (N) \\
2. Ratio of digits to N \\
3. Ratio of letters to N \\
4. Ratio of uppercase letters to N \\
5. Ratio of tabs to N \\
6. Frequency of each alphabet (A-Z), ignoring case (26 features) \\
7. Frequency of special characters: \textless\textgreater\%\textbar\{\} []/$\backslash$@\#\~\ +-*=\$\^\ \&\_()' (24 features). \\
\end{tabular}
\begin{tabular}{|p{8cm}}
\hline
\textbf{Lexical Features - Word-Level}\\\hline
1. Tokens count (T)\\
2. Average sentence length (in characters)\\
3. Average word length (in characters)\\
4. Ratio of alphabets to N\\
5. Ratio of short words to T (a short word has a length of 3 characters or less)\\
6. Ratio of words length to T. Example: 20\% of the words are 7 characters long. (20 features)\\
7. Ratio of word types (the vocabulary set) to T\\
\end{tabular}\\
\begin{tabular}{p{15.5cm}}
\hline
\textbf{Syntactic Features}  \\\hline
1. Frequency of Punctuation: , . ? ! : ; ' " (8 features) \\
2. Frequency of each function words \citep{OShea.J:2013} (277 features)\\
\hline
\end{tabular}
\caption{List of stylometric features.\label{tbl:features}}
\end{table*}

\section{Optimal Hyperparameters\label{appOptimal}}
\begin{table*}[htbp]
\centering
\begin{tabular}{c|l }
\hline
\textbf{Hyperparameter}  &  \textbf{Range}  \\ \hline
$k$ & 100, 200, 300, 400, 500, 1000, 2000, 3000, 4000, 5000 \\
$f_{t} $ & 5, 10, 15, 20, 25, 30, 35, 40, 45, 50 \\
$n_{ch}$ & 3, 4, 5, 6, 7, 8 \\
$n_{w}$ & 1, 2, 3 \\
$epochs$ & 2, 5 \\
$vocab\_size$ & 2000, 5000 \\
\hline
\end{tabular}
\caption{\label{hypers}
Hyperparameters for masking and $n$-gram based feature representations. $k$ is the threshold for masking, $n_{w}$ is the word-level and POS $n$-grams, $n_{ch}$ is the character-level $n$-gram, and $f_{t}$ is the minimum frequency threshold in the whole dataset.}
\end{table*}

See Tables~\ref{hypers} and~\ref{optimalTable}. For FS BERT and RoBERTa, we used the pretrained sequence classification models. These pretrained models do not have hyperparameters for the model structure, but only have pretrained configurations. We used the base uncased models, where base refers to the models' size (not large, and not distilled) and trained on all-lower-case text. For the training procedure, we used the following: AdamOptimizer, lr=0.1, Epochs=500, EarlyStopping(min\_delta=1e-3, patience=100). Despite the large Epoch value, most models would stop after less than 150 epochs.

We implemented~\citet{Barlas2020} ourselves. The code was made available online in a later version~\citet{Barlas2021}. We performed a gridsearch hyperparameter tuning for the number of epochs and the vocabulary size. For the topic confusion task, we used epochs=2 and vocab\_size=2000 based on the ablation studies on Bert reported in~\citet{Barlas2020}. 

\begin{table*}[htbp]
\centering
\begin{tabular}{l|ccrr}
\hline
\textbf{Method}   & \textbf{$k$} & \textbf{$n$} & \textbf{$f_{t}$} & Feat. \\
\hline
Masking (W.)                  & 1,616.7  & 1.9 & 7.9 & 3,265.8 \\ 
Masking (Ch.)                 & 1,691.7  & 5.5 & 18.8 & 6,416.3 \\ 
Stylometric + POS                      & -        & 1.3 & 31.3 & 484.2 \\ 
Stylometric + POS + $n$-grams (W.)  & -    & 2.0 & 12.5 & 2,481.0\\ 
Stylometric + POS + $n$-grams (Ch.) &   -      & 3.8 & 38.3 & 5,355.6\\ 
\hline
\end{tabular}
\caption{\label{optimalTable}
The average optimal parameters for each feature representation, with the resulting number of features under these settings ($k$: masking threshold, $n$: number of tokens in $n$-grams, \textit{$f_{t}$}: minimum frequency threshold in the dataset, W.: word-level, Ch.: character-level).}
\end{table*}


\counterwithout{subsection}{section}
\newpage
\section{Additional Experiments\label{additional}}
\subsection{Data Splitting and Preprocessing\label{sec:prepro}}

\paragraph{The Cross-Topic Scenario.} In all our experiments, we split the dataset into training, validation and test sets. For the cross-topic experiments we followed the same setup in \citep{Stamatatos.E:2017}. We used one topic for training, another topic for validation and hyperparameter tuning, and the remaining two topics for testing. The number of articles was 127 articles when training on Society and 130 articles otherwise. This setup resulted in 12 different topics permutations. We reported the average overall accuracy on all the 12 configurations.

\paragraph{The Same-Topic Scenario.} We combined the 508 articles from all the topics, then split them as follows: 26\% for training, 26\% for validation, and the remaining 58\% for testing. This corresponds to 132 articles for training, 132 articles for validation, and 244 articles for testing. This ensures that the difference in performance between the same-topic and the cross-topic scenarios is not caused by the difference in the number of samples that are used for training/testing. We repeated this process 12 times and reported the average overall accuracy.

\subsection{Cross-Topic Authorship Attribution \label{sec:cross}}
As shown in Table~\ref{classification}, by combining the stylometric features and POS tags with $n$-gram features we achieve the highest accuracy of $83.3\%$. This is in line with our findings in the topic confusion task in Sec.~\ref{sec:confTask}. The difference between using all the features ($mean=83.26$, $SD=2.63$) and the character-based masking approach ($mean=80.89$, $SD=2.59$) is statistically significant ($P=0.04$)\footnote{We used a t-Test: Two-Sample Assuming Unequal Variances at the $\alpha = 0.5$ level.}. 

\begin{table*}[ht]
\parbox{.45\linewidth}{
\centering
\begin{tabular}{l|c|}
\hline
\textbf{Features}   & \textbf{Accuracy} \\
\hline
Stylo. + POS        & 79.2  $\pm$ (2.7)\\ 
Stylo. + POS +  $n$-grams (W.)   &  \textbf{83.3}  $\pm$ (2.6)\\ 
Stylo. + POS +  $n$-grams (Ch.)  &   82.8 $\pm$ (2.7)\\
Masking (W.)      & 77.9  $\pm$ (4.0)\\ 
Masking (Ch.)     & 80.9  $\pm$ (2.6)\\ 
Masking (W.) + Stylo. + POS      & 82.8  $\pm$ (3.3)\\
Masking (Ch.) + Stylo. + POS     & 83.2  $\pm$ (3.3)\\ 
FS BERT  & 37.5 $\pm$ (3.5)\\
BERT AP (K) & 67.3 $\pm$ (4.4)\\
BERT AP (U) & 71.1 $\pm$ (3.3)\\
FS RoBERTa  & 51.1 $\pm$ (3.4)\\
RoBERTa AP (K) & 70.8 $\pm$ (2.0)\\
RoBERTa AP (U) & 75.8 $\pm$ (3.8)\\
\hline
\end{tabular}
\caption{Average cross-topic classification accuracy (\%) on the extended Guardian dataset (W.: word-level, Ch.: character-level).\label{classification}}}
~~~~
\parbox{.45\linewidth}{
\begin{tabular}{|l|c}
\hline 
\textbf{Features} & \textbf{Accuracy}\\
\hline 
Stylo. &  61.2 $\pm$ (3.1)\\
POS & 71.0 $\pm$ (3.2)\\
W. $n$-grams & 77.7 $\pm$ (2.7) \\ 
Ch. $n$-grams & 77.3 $\pm$ (2.8) \\ 
Stylo. $+$ POS   &  79.2 $\pm$ (2.7) \\
Stylo. $+$ POS $+$ $n$-gram (W.) & \textbf{83.2} $\pm$ (2.6) \\
\phantom{Masking (Ch.) + Stylo. + POS}  & \\
\\\\\\\\\\\\
\hline
\end{tabular}
\caption{\label{ablatioin}
Ablation study: classification accuracy ($\%$) on cross-topic scenario. (Stylo.: Stylometric, W.: word-level)}}
\end{table*}

It is also worth noting that by using only stylometric features with POS $n$-grams we can achieve similar results to the masking approach with character-level tokenization. The difference of 1.7\% in favor of the masking approach is statistically insignificant ($P=0.15$) with a ($mean=80.89$, $SD=2.59$) for masking versus a ($mean=79.22$, $SD=2.70$) when using stylometric features with POS $n$-grams. 

Furthermore, Table~\ref{classification} shows a 3\% increase in the accuracy for the masking approach when using character-level tokenization. This outcome is in line with the findings in \citep{Stamatatos.E:2017}. The difference between word-level $n$-grams ($mean=77.90$, $SD=4.03$) and character-level ($mean=80.89$, $SD=2.59$) is statistically insignificant ($P=0.05$). Similarly, the difference between combining the stylometric features and POS-grams with word-level $n$-grams ($mean=83.26$, $SD=2.63$) versus with character-level $n$-grams ($mean=82.83$, $SD=2.7$) is statistically insignificant ($P=0.71$).

Finally, the difference between the state-of-the-art approach which is masking on the character-level from one side, versus stylometric features and POS tags combined with either character-level $n$-grams ($mean=80.89$, $SD=2.59$), masking on the word-level ($mean=82.80$, $SD=3.34$) or masking on the character-level ($mean=83.17$, $SD=3.33$) is statistically insignificant ($P=0.10$, $P=0.98$, and $P=0.80$, respectively.). The only statistically significant difference ($P=0.0.4$) was with stylometric features and POS tags combined with word-level $n$-grams ($mean=83.26$, $SD=2.63$)

\subsection{Ablation Study on the Cross-Topic Scenario\label{sec:abl}}
We conclude our experiments with an ablation study to see the contribution of each set of features to the overall accuracy. Similar to the experiments above, we perform a grid search over all the hyperparameters $f_t$ and $n$. As shown in Table~\ref{ablatioin}, each feature set on its own does not achieve the same performance as with combining all of them. We also confirm the previous results where, even in the cross-topic scenario, $n$-grams outperformed stylometric features by a large margin. We evaluated the significance of the difference between the top three accuracy groups. The results show that the difference between Set (3) ($mean=77.7$, $SD=2.69$) and Set (4) ($mean=79.3$, $SD=2.7$) is statistically insignificant ($P=0.21$) while it is significant ($P<0.01$) between Set (4) and Set (5) ($mean=83.3$, $SD=2.6$). 

\end{document}